\documentclass[conference]{IEEEtran}
\IEEEoverridecommandlockouts
\usepackage{cite}
\usepackage{amsmath,amssymb,amsfonts}
\usepackage{algorithmic}
\usepackage{graphicx}
\usepackage{textcomp}
\usepackage{xcolor}
\def\BibTeX{{\rm B\kern-.05em{\sc i\kern-.025em b}\kern-.08em
    T\kern-.1667em\lower.7ex\hbox{E}\kern-.125emX}}
\begin{document}

\title{Event-based Timestamp Image Encoding Network for Human Action Recognition and Anticipation 
}

\author{\IEEEauthorblockN{Chaoxing Huang}\thanks{This work was done when Chaoxing was at ANU}
\IEEEauthorblockA{\textit{Australian National University} \\
Canberra, Australia \\
chaoxinghuang96@gmail.com}
}

\maketitle

\begin{abstract}
Event camera is an asynchronous, high frequency vision sensor with low power consumption, which is suitable for human action understanding task. It is vital to encode the spatial-temporal information of event data properly and use standard computer vision tool to learn from the data. In this work, we propose a timestamp image encoding 2D network, which takes the encoded spatial-temporal images with polarity information  of the event data as input and output the action label. In addition, we propose a future timestamp image generator to generate future action information to aid the model to anticipate the human action when the action is not completed.  Experiment results show that our method can achieve the same level of performance as those RGB-based benchmarks on real world action recognition, and also achieve the state of the art (SOTA) result on gesture recognition. Our future timestamp image generating model can effectively improve the prediction accuracy when the action is not completed. We also provide insight discussion on the importance of motion and appearance information in action recognition and anticipation.
\end{abstract}

\begin{IEEEkeywords}
event camera, action recognition, action anticipation, spatial-temporal encoding
\end{IEEEkeywords}

\section{Introduction}
Action recognition and anticipation are playing significant roles in multimedia and robotic vision, such as sports video classification\cite{1}, autonomous driving\cite{2}, human robot interaction\cite{3} and safety surveillance\cite{4}. Conventional vision sensor in these tasks is camera, which is common and familiarized in daily life. However, traditional camera suffers from the problems of information redundant, power consuming and motion blurry\cite{5}\cite{6}. Privacy problem is also raised since camera captures a completed scene of the environment, which might include facial and clothing information. Some of the above problems become significant in action recognition and anticipation when the static background is sophisticated or the action movement takes place at a high speed. It also creates power burden for the robot when the robot needs to be constantly deployed without human intervention. In recent years, event camera, a neuromorphic and asynchronous sensor is studied to replace camera in various computer vision tasks, since the attributes of event camera can address the drawbacks of camera which are mentioned above\cite{7}. On the other hand, the raw data representation of event camera is different from conventional camera, thus it is worth looking into the problem of connecting event-based vision and traditional computer vision and use existed powerful tools in action  recognition and anticipation to solve the problem. 
\par{Action recognition stands for recognizing the human action by observing a completed sequence of video\cite{14}. The inference of the label takes place after the action is finished. The action recognition can be further split into forming action representation and action classification. The occurrence of deep neural network (DNN) leverage these two sub-tasks to a certain extend due to  deep neural network's ability of learning complex representation\cite{15}. }
\par{ Action anticipation means inferring the label of human action by observing incomplete video sequence. It can be considered as a general case of action recognition as the algorithm needs to recognize the human action before the action is finished\cite{15}. Therefore,human action anticipation is also referred to early action prediction. The difference of human action recognition and anticipation is shown in Fig.1.
}
\begin{figure}[ht]

\centering
\includegraphics[scale=0.25]{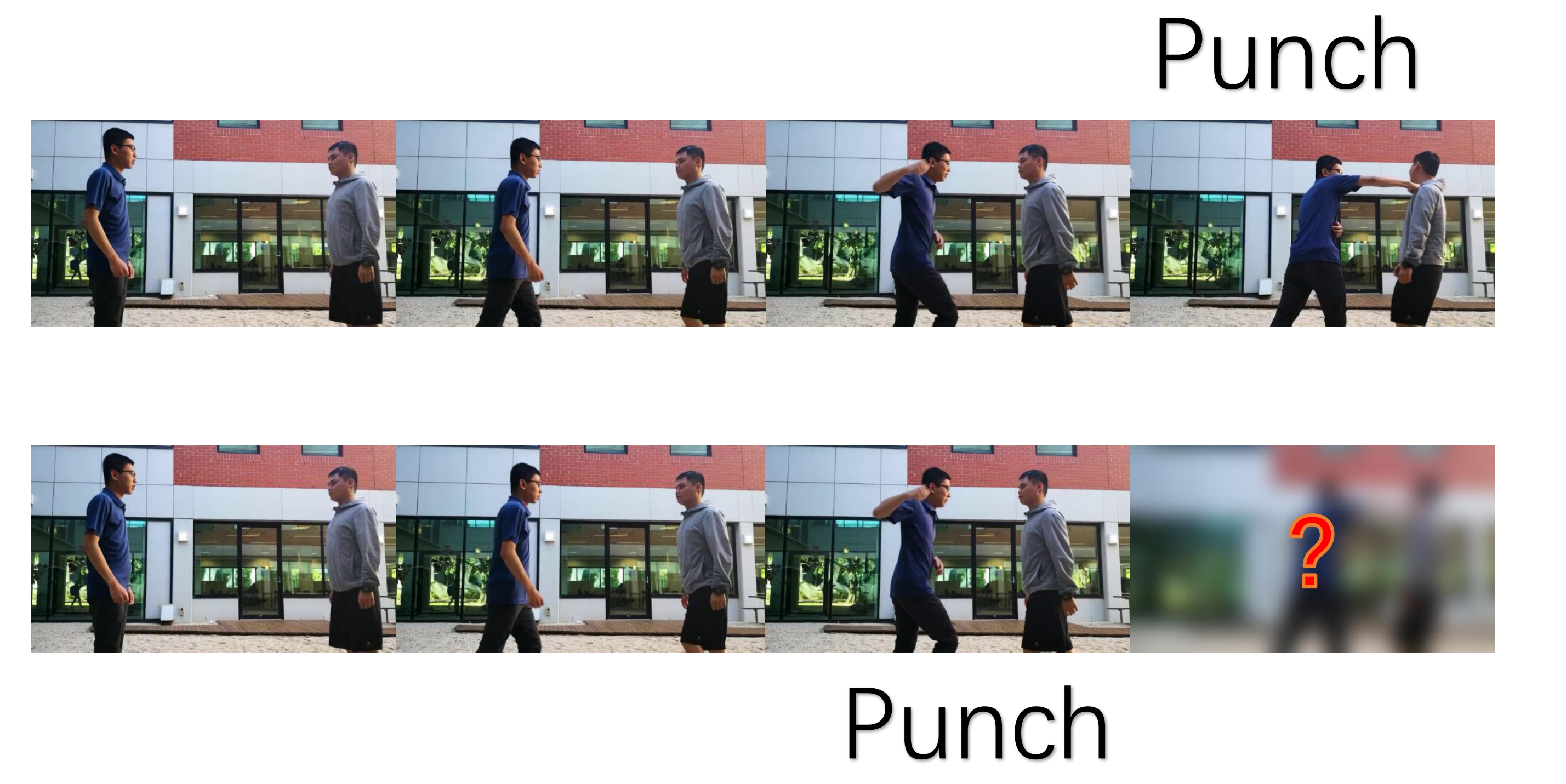}
\caption[Difference between action recognition and anticipation]{Difference between action recognition and anticipation.The sequence in the first row is action recognition and the sequence in the second row is action anticipation.}
\label{fig:label}
\end{figure}
 \par Few works have connected deep learning method with event-based human action recognition and anticipation. Amir et al. \cite{12} use event image to encode the spatial information into an image and apply it to the 2D CNN for hand-gesture recognition,which ignores the temporal information. Wang et al. \cite{29} encode event data point into spatial temporal point-cloud, while it does not reflect the polarity of the event. 3D CNN is used in \cite{41} to perform RGB-based action recognition, which introduce a large number of parameters. The work in \cite{20} uses timestamp image to encode the spatial-temporal information in event data for human gait recognition, which is similar to our method.  However, they do not consider the polarity of the event. Besides, the model in those work are only applied to simple and single scenario in action recognition like hand gesture recognition and gait recognition.

To echo the motivation and the aforementioned problems, we propose a method of using polarity based timestamp image encoding to encode the spatial-temporal information as well as polarity information from the event data sequence into frame-based representation, and use standard 2D CNN to classify the action label of the observed event sequence. The contributions of this paper are: \textbf{1)} We propose a simple but compact CNN-based method to perform event-based action recognition by using the timestamp image to learn the  spatial-temporal as well as polarity information to aid the action understanding. \textbf{2)} We compare the performance of applying purely spatial encoding (event image representation) as well as spatial-temporal encoding (timestamp image representation), and demonstrate that temporal information is indeed useful in event-based action recognition. \textbf{3)} We demonstrate that our event-based recognition method can achieve a competitive performance result on real world action scenario compared with those RGB-based method, even without using the RNN/LSTM \cite{lstm} architecture to exploit temporal dependency. Our recognition method also surpass the  state of the art in gesture recognition task. \textbf{4)} To our knowledge, we are the first to look into the problem of event-based action anticipation.  Compared with purely action recognition, we demonstrate that by generating future class-oriented timestamp images, the prediction performance of inferring action class can be improved during the early stage of the action.

\section{Related works}
The development of RGB-based action recognition and anticipation is rather mature and we refer readers to \cite{14} for details. We pay emphasis on the representation of event-data, since this is the main gap between event-based vision and standard learning-based computer vision. 
\subsection{Representation of event data}
The raw data from event camera is nothing but just a sequence of asynchronous tuples. Thus it is essential to transform those data into features that can be utilised by computer vision method. Inspired by frame-based computer vision, the most intuitive representation construction method is projecting those event to the frame-plane. One method that is widely explored is by encoding the timestamp value into pixel intensity value, which is called the time-surface \cite{18,19} or timestamp image\cite{18}. Wang et.al \cite{20} use timestamp image as input of a CNN to recognize the gaits, which is similar to action recognition that requires the model to capture motion information. Park et.al \cite{21} use timestamp image in spatial-temporal demosaicing for gesture recognition. The above works do encode the spatial-temporal in a visually intuitive way, but ignore the polarity information, which may reflect the direction of the motion since motion at different direction can cause different brightness changes. Another frame-based representation is called the 2-D histogram of count or event-image\cite{22}. Works\cite{23,24} show that event-image is closely connected with moving edges and event-image representation is generally adopted in mobile robotics such as vision control and localization\cite{25,26,27}, since edges of the environment matter in these tasks.  Zhu et.al\cite{18} combine both timestamp image and event-image for end to end estimating optical flow. However,it should be pointed out that event-image does not include the temporal information and the visual appearance only includes those moving edges. Zhu et.al \cite{28} also proposed a 3D-voxel event-representaion, which is no longer a frame-based, for unsupervised optical-flow estimation. The above two works on optical flow estimation further solidify the statement that event-based data can provide motion information, which is aligned with the requirement of action understanding. In \cite{29}, the events are novelty presented as 3D point-cloud  to further reduce the redundancy of frame-based event representation, and a PointNet++\cite{30} method was used for gesture recognition. However, the polarity is not included in this kind of representation.  Recently, deep learning approach has been adopted for end to end event representation learning\cite{31} and there are works directly transform event data into real camera frame to leverage the bridge between event-based vision and traditional computer vision\cite{32,33}.

\section{Methodology}
\label{sec:pagestyle}
\subsection{Event to Frame Transformation}
We first briefly review the the model of event-camera \cite{10}. Let us define the intensity of brightness of an independent pixel as $I$, and what the hardware of the photoelectric system precepts is the log photocurrent, which is denoted as $L=\log(I)$ (We refer it as “intensity” in the following context). We also define the information that the pixel collects as "event", which is denoted as $e=((x,y),t,p)$. $(x,y)$ is the position coordinate of the pixel in the sensing matrix, $t$ is the timestamp when the event is logged by the pixel and $p$ is the polarity, which indicates if the intensity at this pixel  becomes brighter or darker. The intensity change is determined by the hardware threshold. The polarity can be further expressed as $p=sign(\frac{d}{dt}\log(I_t(x,y)))$, where the output of $sign$ function is $+1$ or $-1$. In reality, multiple pixels in the sensing matrix record their events, and we get sequence of event data. 

 To bridge the gap between CNN and event data, two methods of projecting events into frame-based representation are presented below,namely timestamp-image and event-image.
 \subsubsection{Timestamp image}
 Timestamp image mainly encodes the time information of the event data into image form. Given a sequence of event data, we consider using a sliding-window with fixed time-length to accumulate the events until the difference between the latest timestamp value and the earliest timestamp value equals to the window length. The pixel value of every pixel is determined by the relative time value within the time-window. The mathematical model is given as below:\\
 Without lost of generality, we assume  $n$ events in  pixel $(x,y)$ take place within the time-window, which are $[e_1(x,y,t_1,p_1),\\ \cdots,e_n(x,y,t_n,p_1)]$ (ascending sorted according to time). We also have the latest and earliest timestamp values as $t_{end}$ and $t_{begin}$ within the time-window by taking all the events into consideration. The pixel value at $(x,y)$ can then be expressed as:
\begin{equation}
I_{x,y}=\frac{t_n-t_{begin}}{t_{end}-t_{begin}}
\end{equation}
Equation (1) indicates the latest relevant time of the event takes place in the given time-window.
\subsubsection{Event image}
The event-image is also called 2D-histogram, since it encodes the amount of event that take place in a certain pixel into the corresponding pixel value. The pixel value at $(x,y)$ can then be expressed as:\\
\begin{equation}
    I_{x,y}=\sum_{t \in T}\delta(x_k=x,y_k=y)
\end{equation}
where $T$ is the length of the time-window and $\delta(\cdot) $ stands for Dirac function.  Equation (2) indicates counting the number of event at every pixel.
\subsubsection{Encode the polarity}
Polarity value(+1 or -1) does not appear in the above two equations, but this information cannot be simply ignored in event-based vision, since the direction of the movement cause different intensity changes. To do that, we simply apply the timestamp image and event image transformation to both positive and negative event separately, which means  two frame-representations will be acquired, namely $I_p$ and $I_n$. We then merge these two frames into one frame representation as in \cite{26}, where the pixel values in $I_p$ is positive and the pixel values in $I_n$ is negative.  All the pixel values are re-scaled to the range from 0 to 255 after merging. The visualisation of both representations are shown in Fig. 2

\begin{figure}[t]

\centering
\includegraphics[scale=0.35]{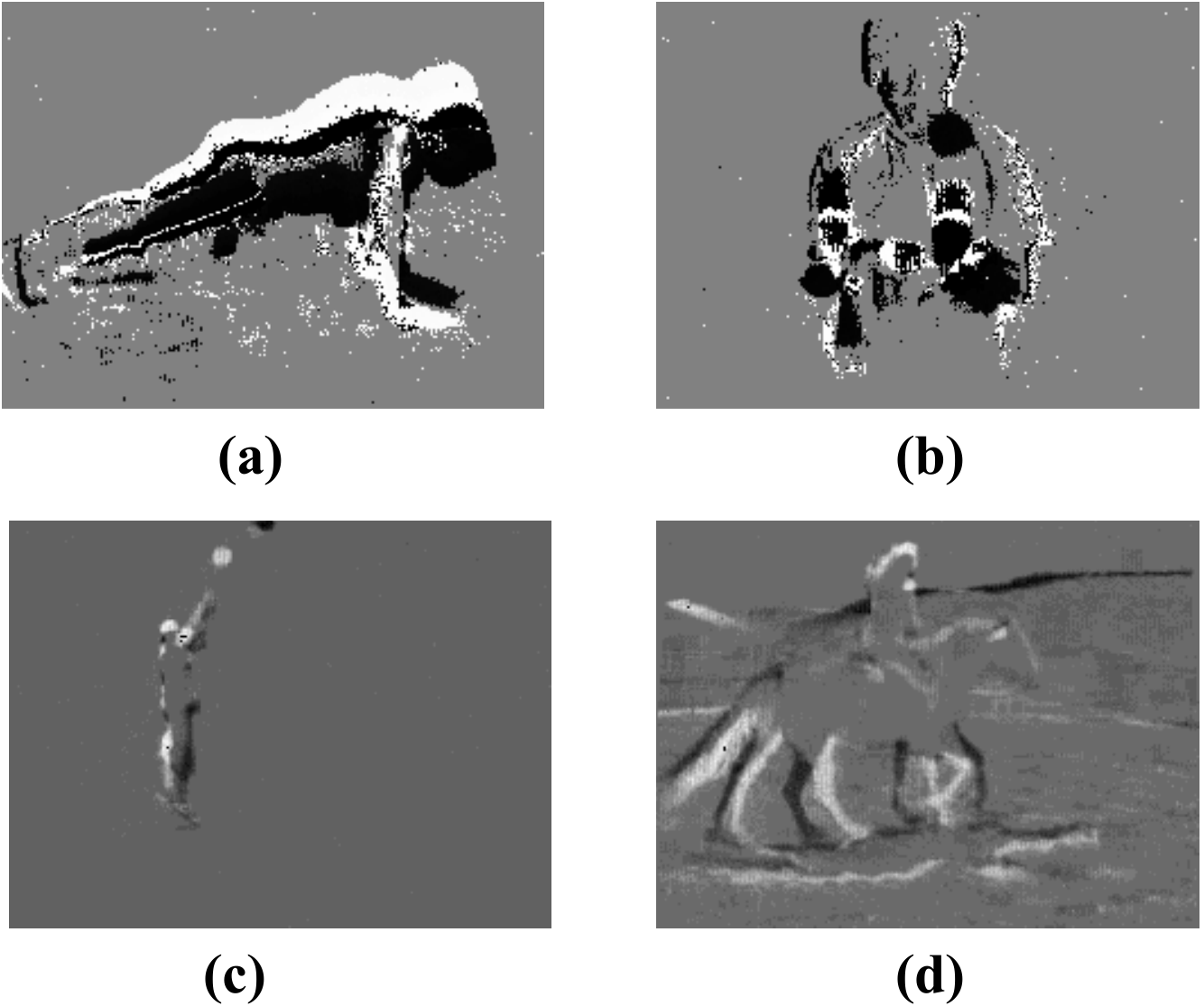}
\caption[Examples of timestamp image and event image]{Examples of timestamp image and event image.(a) Timestamp image of push up.(b) Timestamp image of ball juggling. (c) Event image of basketball throwing. (d) Event image of horse-riding. }
\label{fig:label}
\end{figure}

\subsection{Action recognition}
\subsubsection{Timestamp-image chunk classifier }
We transform the event data into frame representation in the previous section, and we can now use CNN to apply recognition task to image data. We do not use RNN/LSTM architecture like RGB-based vision since we want to demonstrate the  temporal information encoded in 2D layout is sufficient for action understanding without explicitly learning the temporal dependency, and to utilise the advantages of event camera. One might consider training a classifier to recognize every frame, but it is problematic that some of the frames record very sparse pixel value since event camera is only sensitive to those moving objects. Those frames tend to have the moving background and some of the noisy pixels, which yields difficulties in recognizing human and other objects in the scene. If we directly train the classifier on a single frame, those noisy frames will create severe over-fitting problem and the classifier might even learn wrong features. To overcome this problem, we use a sliding frame-buffer with a step size of one frame, and it stores three event frame representation at every step. The stored three frames is referred as timestamp-image chunk, as we use timestamp-image representation in our model. Note we do not  consider larger chunk size because we want to show the effectiveness of spatial-temporal encoding by avoid sequencing much of the data. At this stage, we train a classifier to classify the action label of every chunk. In other words, the classifier is trained to classify every single frame by looking two frames backward. The process of forming timestamp-image chunk is shown in Fig. 3. Following the network architecture in \cite{64}, we use the Pytorch version Resnext50 \cite{66} with pre-trained weight on Image-net as our classifier.

\begin{figure}[t]

\centering
\includegraphics[scale=0.9]{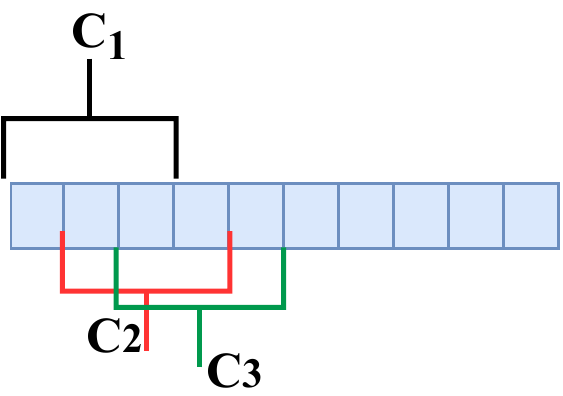}
\caption[Timestamp image chunk]{Process of creating timestamp image chunk. The buffer rolls one frame ahead as the video process.}
\label{fig:label}
\end{figure}
\subsubsection{Inference}
\par{When inferring the action label of an event video which has been transformed into frame representation, we will first use the sliding buffer to create chunks. Every chunk will be sent into the trained chunk-classifier and a class score vector will be obtained for every chunk.  Finally, temporal average pooling is applied to all the score vectors to get the final class score vector of the frame-based event video. The inference pipeline is shown in Fig.4}

\begin{figure}[t!]

\centering
\includegraphics[scale=0.4]{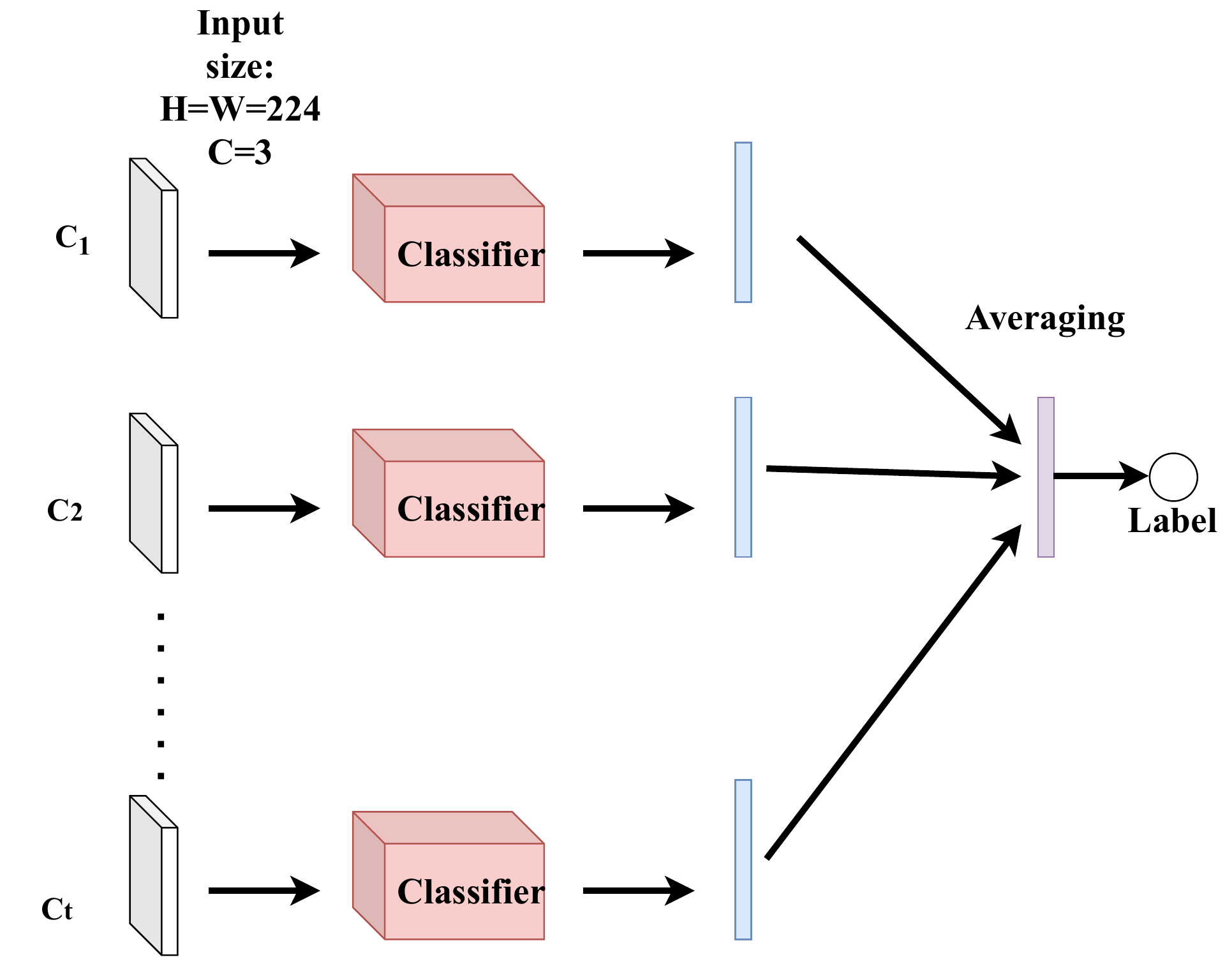}
\caption[Inference in action recognition]{Inference in action recognition}
\label{fig:label}
\end{figure}

\subsection{Action anticipation}
\subsubsection{Observation Ratio}
The event video stream from event camera is now transformed into frame representations, and it becomes consistent to apply the concept of observation ratio which is originated in the action anticipation task in RGB frame-based video \cite{65}.  The observation ratio defines the progress of the action video, which is the complete extent metric of an action in action anticipation. Given a complete video with $N$ frames, and the video is segmented into $S=10$ segments, then each segment of the video contains $\frac{N}{S}$ frames, where $N$ varies from video to video. We will observe $n\times \frac{N}{S} $ frames of video if we observe $n$ segments of the video. The observation ratio is then expressed as $r=\frac{n}{N}$. We will evaluate the recognition accuracy at video level under different observation ratio in action anticipation task. Since action anticipation is a general case of action recognition, the observation ratio in action recognition can be defined as 1. 
\subsubsection{Future timestamp image generator}
In action anticipation, it is essential for the model to "know" the future motion or dynamical information.Inspired by \cite{17}, we denote the current chunk at frame $t$ as $C_t$ and a future chunk of timestamp-image as $C_{t+n}$,  where  $n$ is a positive integer. Chunk $C_t$ is composed with the frame from $I_{t-2}$ to the current frame $I_t$  and chunk $C_{t+n}$ is composed with the future frame from $I_{t+n-2}$ to the future frame $I_{t+n}$. Our goal is to generate future chunk by using a parametric conditional probabilistic model by taking the historical chunks into consideration,which can be expressed as,
\begin{equation}
    P(C_{t+n}|[C_{t},C_{t-1},\cdots],\theta)
\end{equation}
where $\theta$ is the model parameters. The reason that we   predict chunk $C_{t+n}$ directly from the history till $C_t$ rather than using historical information till $C_{t+n-1}$ is because every two adjacent chunks have two frames of overlapping and it can be computational redundant if we need to recursively infer all the future chunks from $t$ on until we get $C_{t+n}$. To further simplify the model, we only takes the current chunk of frame representation $C_t$ as historical information to generate the predicted chunk. The model becomes 
\begin{equation}
    P(C_{t+n}|[C_{t},C_{t-1}],\theta)=P(C_{t+n}|C_{t},\theta)
\end{equation}
Since we need to generate a 3D tensor by taking a 3D tensor as input, a CNN-based auto-encoder is used to be the generator. In real world application of human action recognition and anticipation, the human and the objects that the human interacts with are usually at different scale. This phenomenon is caused by different types of action behaviour or different view perspective. Thus it is crucial to use the information at different scale. Skip connection in CNN-based auto-encoder address this problem\cite{67}, and we use a U-Net\cite{67} architecture with Resnet50\cite{68} as the encoder backbone\cite{69}.
\subsubsection{Training pipeline}
\par{As being mentioned in the action recognition section, we first slice the event video into chunk level and use those chunks to train the classifier network. The training of the classifier is also at chunk level, and the cross-entropy loss function can be expressed as: }
\begin{equation}
    L_{c}=-\sum_{i}y_ilog(f(C))
\end{equation}
where $y_i$ is the ground-truth label of class $i$ and the $f(\cdot)$ is the chunk classifier network function. 
\par{For the training of the future timestamp image generator, we train the network for two objectives. Namely spatial-appearance aspect and categorical aspect. Our first goal is to ensure that the spatial pixel value of the generated future timestamp-image chunk is as close to that of the ground-truth future timestamp-image chunk. In other words, the hallucinated chunk should look like  the real world chunk at the future time-step. The loss function can be expressed as:}
\begin{equation}
    L_s=||C_{t+n}-\hat{C}_{t+n}||
\end{equation}
where $C_{t+n}$ is the hallucinated chunk from the generator and $\hat{C}_{t+n}$ is the ground-truth chunk. We use $L_1$ norm here because it is robust to noisy data and it is adopted here to suppress the effect from those noisy frames with sparse pixel values. 
\par{Apart from enforcing the generated future chunk to be highly similar to the ground-truth from a spatial-appearance perspectives, it is also important that the hallucination representation belongs to the right category. Besides, it is difficult to ensure that the quality of the generated chunk does not degrade the classifier's performance, as the inference process involves the hallucinated chunk. We adopt the teacher-student mechanism being used in \cite{17} to address this problem. We assume the chunk classifier is well-trained and we assign the teacher-role to the classifier to teach the auto-encoder(the student network) to produce categorical-oriented future chunk. At this stage, the weight of the classifier is fixed. The loss function is similar to equation (5), by changing $C$ into $g(C_{t})$, where $g(C_{t})$ is the hallucinated future chunk generated by the auto-encoder. The final loss function for training the future timestamp image generator can then be written as:  }
\begin{equation}
    L=\lambda_1L_c+\lambda_2L_s
\end{equation}
where $\lambda_1$ and $\lambda_2$ are weighted-coefficient. The training process of the auto-encoder is illustrated in Figure 5.

\begin{figure}[t!]

\centering
\includegraphics[scale=0.4]{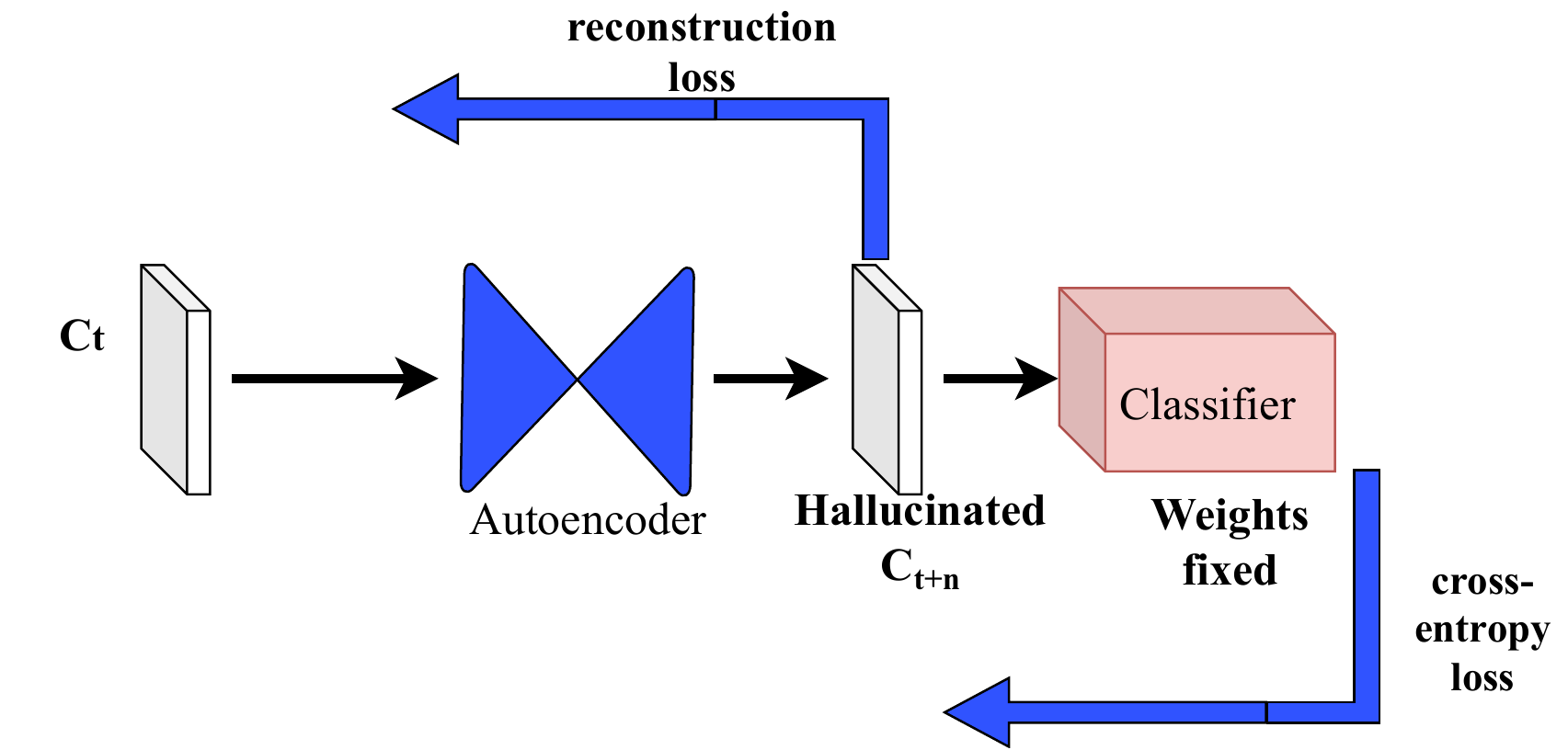}
\caption[Training pipeline of the generator]{Training pipeline of the generator}
\label{fig:label}
\end{figure}
\subsubsection{Inference in action anticipation}
\par{In action anticipation task, we will first use the pipeline in action recognition to obtain the score vectors of the observed chunks. After that, the trained generator takes the final chunk of the observed sequence as input and recursively generate $k$ future chunks. Lastly, temporal average pooling is applied to all the score vectors from observed chunks and hallucinated chunks. The inference pipeline is shown in Fig.6 }
\begin{figure*}[t!]

\centering
\includegraphics[scale=0.5]{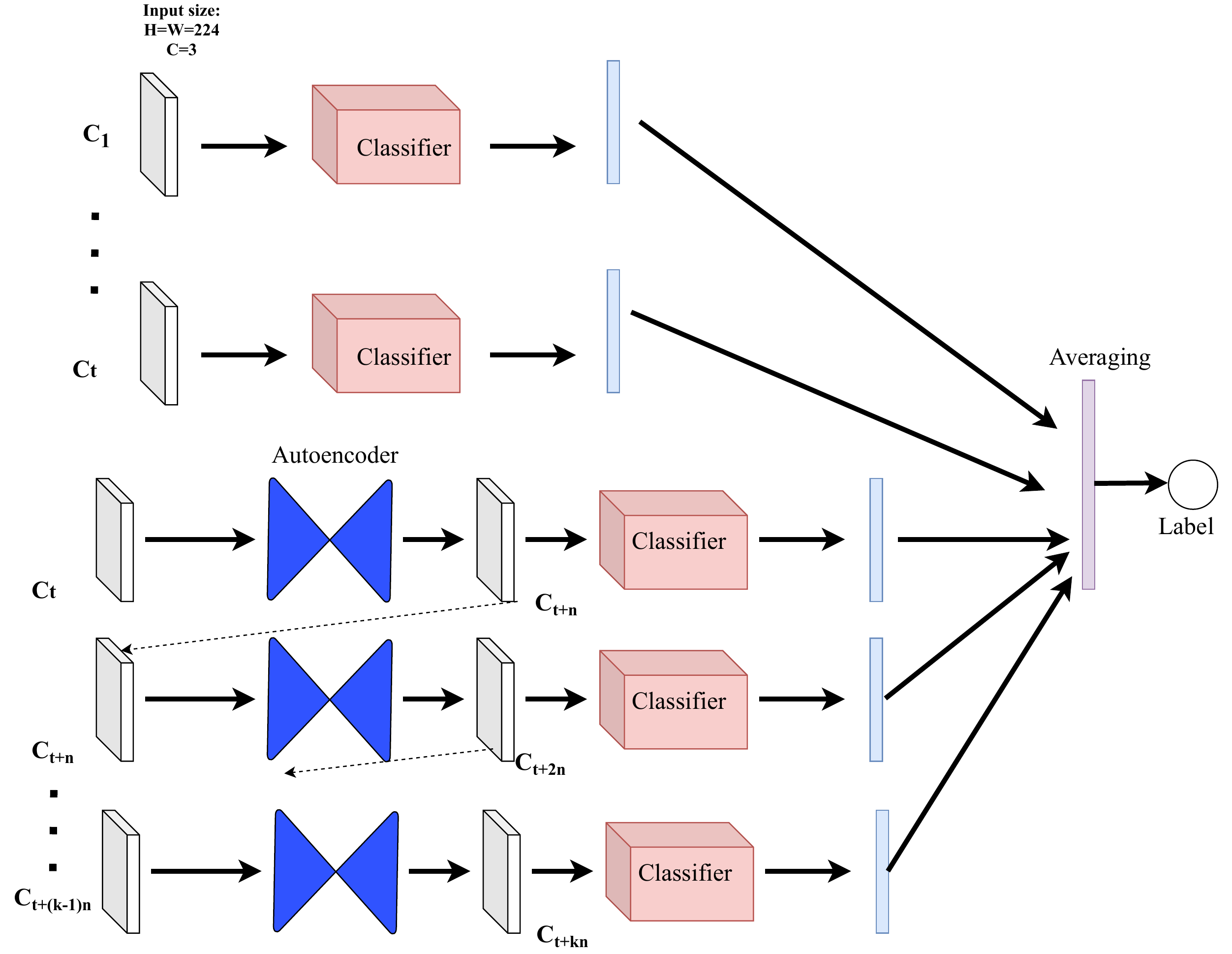}
\caption[Inference in action recognition]{Inference in action anticipation}
\label{fig:label}
\end{figure*}

\section{Experiments and results}
In this section, we demonstrate experiments on two challenge data-set. We will first introduce these two datasets. 
\subsection{Datasets}
 \textbf{DVS-128 gesture dataset } is an event-based human gesture dataset which is produced by IBM to build a low power cost event-based gesture recognition hardware system \cite{12}.  It was recorded by using the iniVation DVS-128 dynamic vision sensor to record a set of 11 hand and arm gestures from 29 subjects with 122 trails under 3 various lighting conditions. The number of the total instances is 1,342.  In our experiment, the gesture of "Other" is omitted and only 10 of the pre-designed gestures are included. We follow the standard train-test split provided by the dataset. 
 
 \textbf{Event-version UCF-11 dataset} is recorded by playing the RGB dataset UCF-11 video on a computer monitor and a DAViS240C vision is put in front of the screen to get the event-data. The position of the vision sensor is adjusted to ensure the field of view can cover the region of interest on the monitor, and the video sequence is re-scaled to fit the field of view of the vision sensor. The video numbers (16000) and the categories remain the same as the RGB version apart from changing the data into event data as AEDAT form. The action samples in the dataset are all from real world videos. We refer the readers to the RGB version dataset for more details \cite{77,37}. The data split and group (fold) forming are strictly aligned with the RGB version. Note that we do not use the UCF-50 dataset due to the 25-fold validation is computationally consuming, while the event-version UCF-101 is not published.
 \subsection{Implementation details}
 The hardware platform are  4 NVIDIA Titan-X GPUs and the i9-9900 CPU. The experiment environment is Ubuntu 16.04 and the framework is Pytorch 1.1.0 \cite{70}. The time-window of collecting event data for generating each frame representation is set as 80 ms and each frame is resized as 224 $\times$ 224. Random flipping and brightness jittering are used for data augmentation. As for the training of the classifier, the optimizer is Adam with $\beta_1=0.9$ and $\beta_2=0.999$. The learning rate and batch size are $1e-4$ and 64 respectively and a regularization of $1e^{-5}$ is used for prevent over-fitting. Turning to the training of the generator, $\lambda_1$ and $\lambda_2$ are set as 1 and 0.5 respectively, other configuration are the same as the training of the classifier. As for the future hallucinated chunk $C_{t+n}$, $n$ is set as 3, which exactly includes the next 3 frames of timestamp images and does not has overlapping with the current chunk.  The number of future chunks $k$ is set as 2. 
 \subsection{Action Recognition}
 \subsubsection{Comparison on different event-representations}
We first compare the performance of using timestamp image chunk and  event-image chunk. For simplicity here, we group the 25 groups of data into 5 larger groups (fold 1 to fold 5 as the first group, fold 6 to fold fold 10 as the second group and so on. ) and do  5-fold validation for both representations respectively. This improves the difficulty  level of the recognition task since the number of learning sample is considerably reduced compared  with the original leave-one out validation(LOOV). Each fold of training takes 20 epochs. The results are shown in Table \uppercase\expandafter{\romannumeral1} . As is shown, the performance of using timestamp-image chunk is higher than that of using event-image chunk, which does not have temporal information. We therefore adopt polarity-based timestamp image as our representation in our following experiments. 
 \begin{table}
 \centering
\caption{Performance comparison of two different representations }
  \setlength{\tabcolsep}{10mm}{
  \begin{tabular}{||c c ||} 
 \hline
 \textbf{Representation} & \textbf{Accuracy(\%)}  \\ [0.5ex] 
 \hline\hline
 Timestamp-chunk & 83.23 \\ 
 \hline
 Event-chunk & 82.00 \\[1ex] 
 \hline
\end{tabular}
}
\end{table}

\subsubsection{Results on EV-UCF-11 dataset }
We conduct the experiment on human action recognition by using the EV-UCF-11 dataset, and we strictly follow the LOOV method to test the performance for consistency because we need to compare it with those benchmarks which use 25-fold validation. Benchmarks that use EV-UCF-11 as well as UCF-11(RGB) dataset are included in the comparison. The results are shown in Table  \uppercase\expandafter{\romannumeral2}. The result indicates that the proposed method outperforms the existing event-version benchmark(Motion map + MBH) significantly and reach the same performance level as those of RGB benchmarks, even it does not have color and background information.

\begin{table}
\centering
\caption{UCF-11 results comparison}
    \setlength{\tabcolsep}{8mm}{
 \begin{tabular}{||c c ||} 
 \hline
 \textbf{Method} & \textbf{Accuracy(\%)}  \\ [0.5ex] 
 \hline\hline
 Motion map + MBH\cite{71} & 75.13 \\ 
 \hline
 Two-stream LSTM \cite{73}& 94.60\\
 \hline
 Differential motion\cite{74} & 90.24\\
 \hline
 CNN+TR-LSTM\cite{75} & 93.80\\
 \hline
 CNN+LSTM\cite{75} & 92.30\\
 \hline
 \textbf{ Proposed} & \textbf{92.90}\\[1ex] 
 \hline
\end{tabular}
}
\end{table}
\subsubsection{Results on DVS-128 gesture dataset}
We also conduct gesture recognition experiments by using the DVS-128 gesture dataset and achieve the state of the art result. The results are shown in Table  \uppercase\expandafter{\romannumeral3} . We also try to use a smaller or larger time-window size for event data collection, which turns out to have a lower performance.
\begin{table}[t!]
\centering
\caption{DVS-128 results comparison}
\setlength{\tabcolsep}{8mm}{    
 \begin{tabular}{||c c ||} 
 \hline
 \textbf{Method} & \textbf{Accuracy(\%)}  \\ [0.5ex] 
 \hline\hline
 Pointnet ++ \cite{29} & 97.06 \\ 
 \hline
 Time cascade \cite{12} & 96.49\\
 \hline
 
 \textbf{Proposed, T=80ms} & \textbf{97.35}\\
 \hline
 \textbf{ Proposed,T=50ms} & \ 95.83\\
 \hline
 \textbf{Proposed, T=20ms} & \ 95.83\\[1ex] 
 \hline
\end{tabular}
}
\end{table}

\subsection{Action Anticipation}

\subsubsection{The effectiveness of generating future chunks}
We first compare the anticipation performance under different observation ratios of purely using the classifier and that of using classifier with generated future chunks. The dataset in this task is the EV-UCF-11 dataset and we conduct the 5-fold-validation as what is done in action recognition.  The results are shown in Fig .8.  Note we do not carry out experiment on DVS-128 dataset because the gestures are periodic actions where action anticipation becomes unnecessary and meaningless. In Fig .8 , the inference performance can be improved after taking the generated future chunks into consideration, especially during the early stage of the video(prior to 50\% of observation ratio). The improvement becomes marginal  in the second half of the video since the observed part is already with high certainty. The generated chunks may even degrade the performance at the late stage of the video (80 \%) due to their accumulated uncertainty. 

\subsubsection{Comparison with RGB benchmark }
We compare  the anticipation of our event-based algorithm with the RGB-based benchmark in \cite{17}. The datasets here are the UCF-11 and EV-UCF-11. In \cite{17}, the experiment is tested on the split-1 of the UCF-101-24 dataset\cite{76}, which contains the UCF-11 dataset. For conventionality, we carry out the test in the RGB-based experiment and event-based experiment on the split-1(group 1 to group 7) of the UCF-11 and EV-UCF-11 dataset respectively. Note the observation ratio here is set as 10\% as in \cite{17}, since it has been shown that there is no significant performance improvement for larger observation ratio\cite{17}. The performances are shown in Table \uppercase\expandafter{\romannumeral4} and the confusion matrix of the RGB and event-based anticipation are shown in Fig.9 and Fig.10 .\\
The results reveals that the performance of the proposed  event-based algorithm is 10\% less than that of the RGB baseline,  which does not have the same level of performance, and we further find out that 3 of the actions contributes the majority of the mis-classification(34/46), namely TennisSwing, GolfSwing and WalkingWithDog. Further discussions will be provided in the next part.
\begin{table}
\caption{Anticipation performance comparison with RGB-based method }
    \centering
    {
 \begin{tabular}{||c c c ||} 
 \hline
  & RGB-based\cite{17} & Proposed  \\ [0.5ex] 
 \hline
  Without anticipation & 89.57\% &79.78\%  \\ 
 \hline

 Anticipation & 90.00\% & 80.00\%  \\ [1ex] 
 \hline
\end{tabular}
}
\end{table}

\subsubsection{What leads to the performance gap ?}
We demonstrate some of the early frame representations of event-based vision and RGB-based vision of those three hard categories in Fig .7. One(human) might find it rather hard to know what the human in the video is doing by only observe the frame at early stage without taking a look at the RGB frame. This drive us to question what really matters in action recognition and anticipation. \\
\par \textbf{Discussion:} 
To start with, motion is considerably helpful in completed action label inferring. This is aligned  with the attribute of event camera, which is sensitive to moving object and mainly records the information of the motion.  Even when background or color information are absent, it is still possible to accurately infer the action label if we witness enough motion information like moving edges. Thus, it is not surprising that event-based action recognition can achieve  a similar level performance as those of RGB-based action recognition model.
\par As for action anticipation, the motion information from event camera may not be enough for producing a good prediction, especially during the early stage of the action. At this stage, the human just start to move and the motion information is not enough. Concretely,  these three kinds of dominated hard examples for our event-based anticipation model relies  heavily on the recognition of the objects that the human interact with and the very first glance of the background or color information of the corresponding RGB frame can already give confidence in identifying the label. One may question why  event-based action recognition can achieve a  similar  level  performance  as  those  of  RGB-based  action recognition model, this is because the complete progress of those moving egdes have been witnessed. 
\begin{figure}[]
\centering
\includegraphics[scale=0.4]{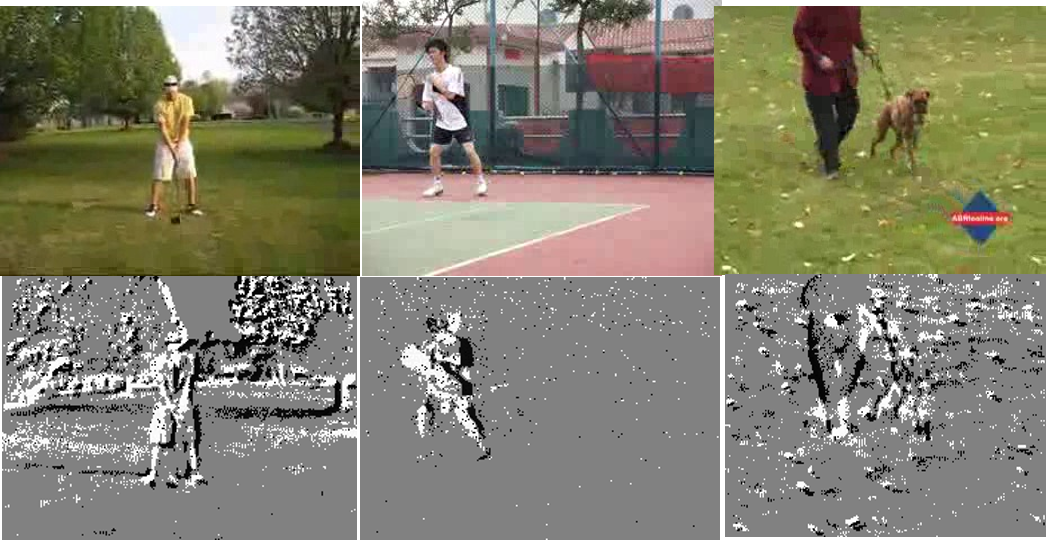}
\caption[Training pipeline of the generator]{Early frames comparison. Golf, Tennis and Walking with dog}
\label{fig:label}
\end{figure}

\begin{figure}[]

\centering
\includegraphics[scale=0.4]{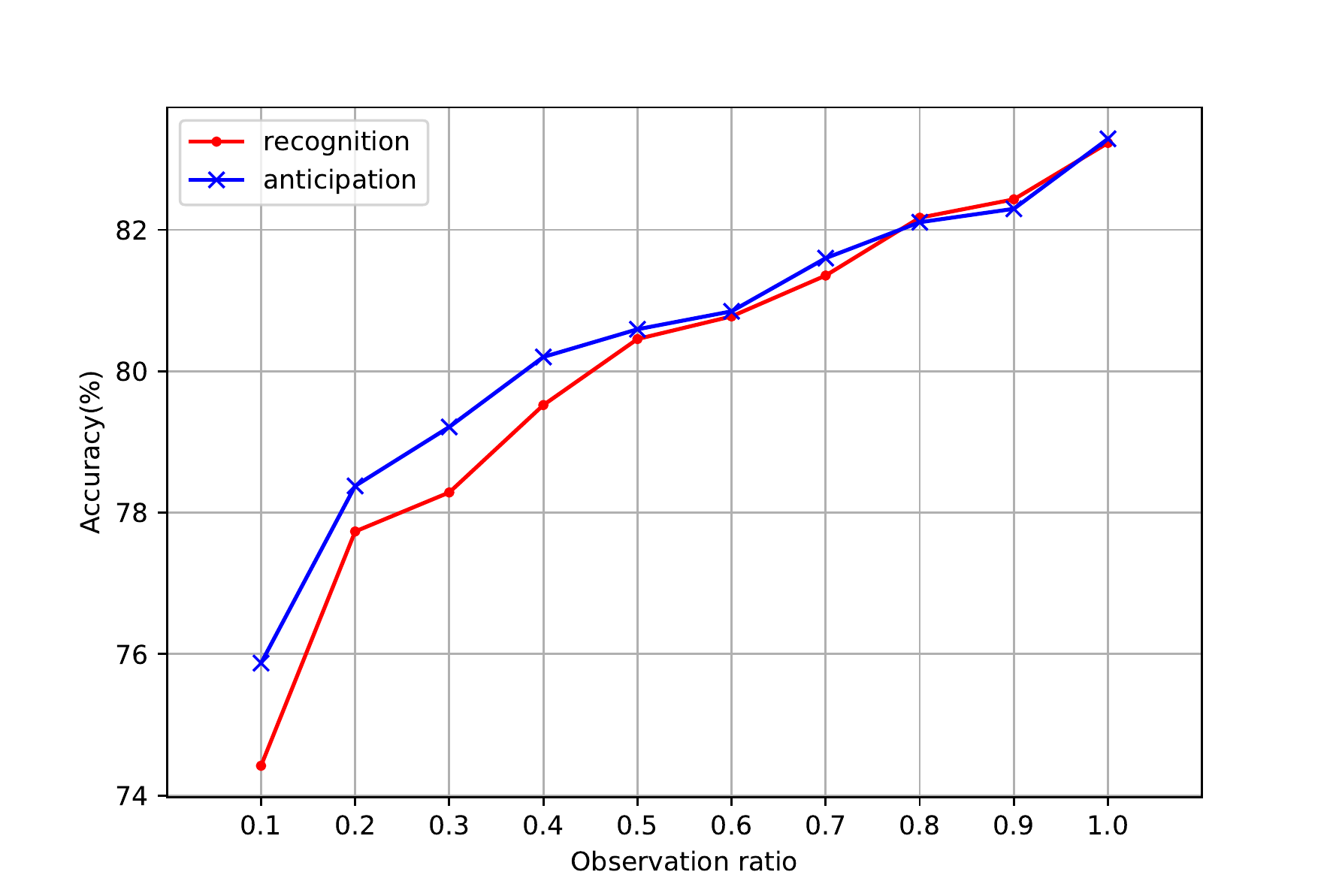}
\caption[]{Inference performance of generating future chunks and pure recognition}
\label{fig:label}
\end{figure}

\begin{figure}[]

\centering
\includegraphics[scale=0.6]{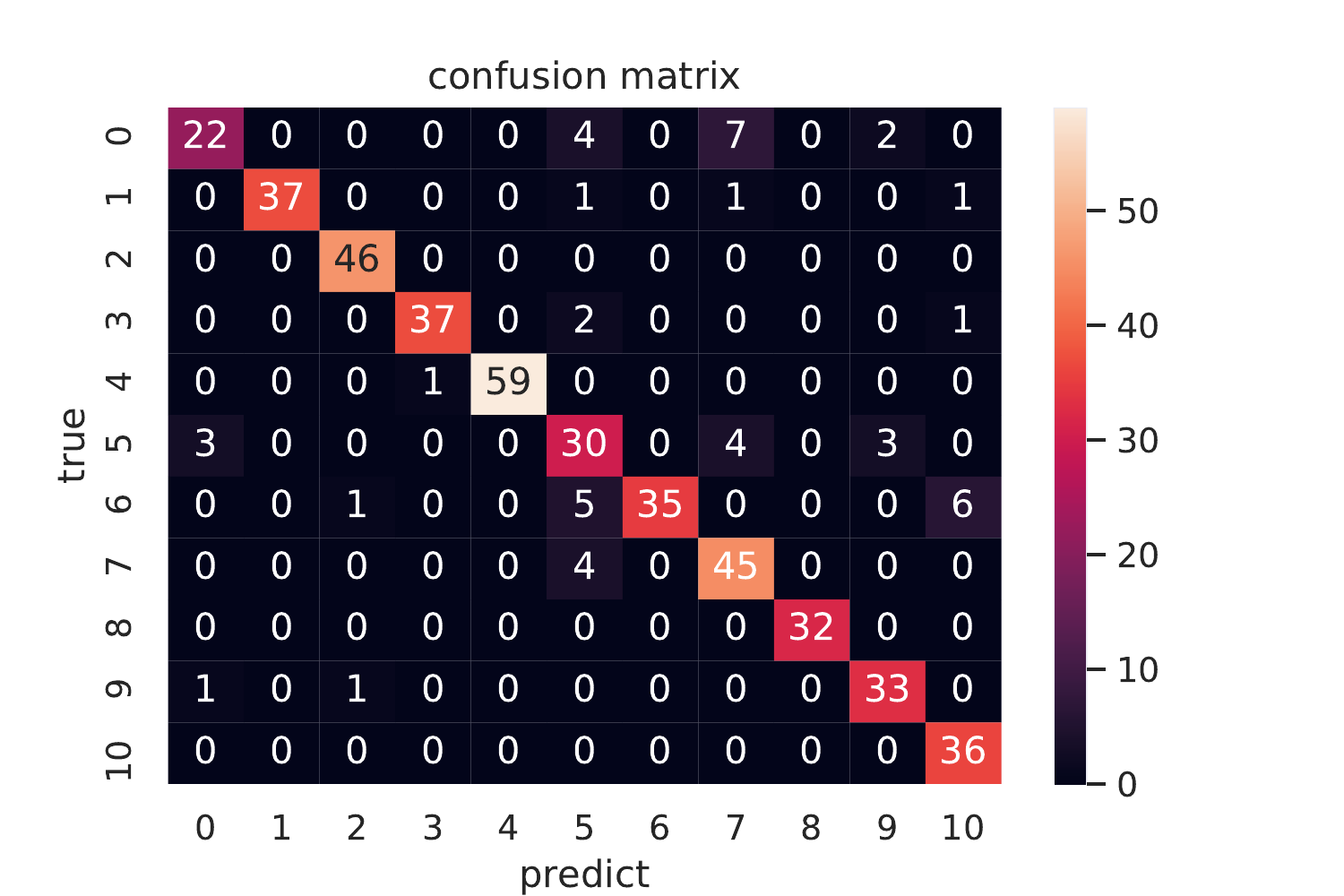}
\caption[Confusion matrix of RGB-based anticipation]{Confusion matrix of RGB-based method. Label dictionary:["Basketball":0, "Biking": 1, "Diving": 2,"GolfSwing":3, "HorseRiding":4, "SoccerJuggling":5, "Swing":6, "TennisSwing":7, "TrampolineJumping":8,"VolleyballSpiking":9,
"WalkingWithDog":10] }
\label{fig:label}
\end{figure}

  \begin{figure}[t!]

\centering
\includegraphics[scale=0.6]{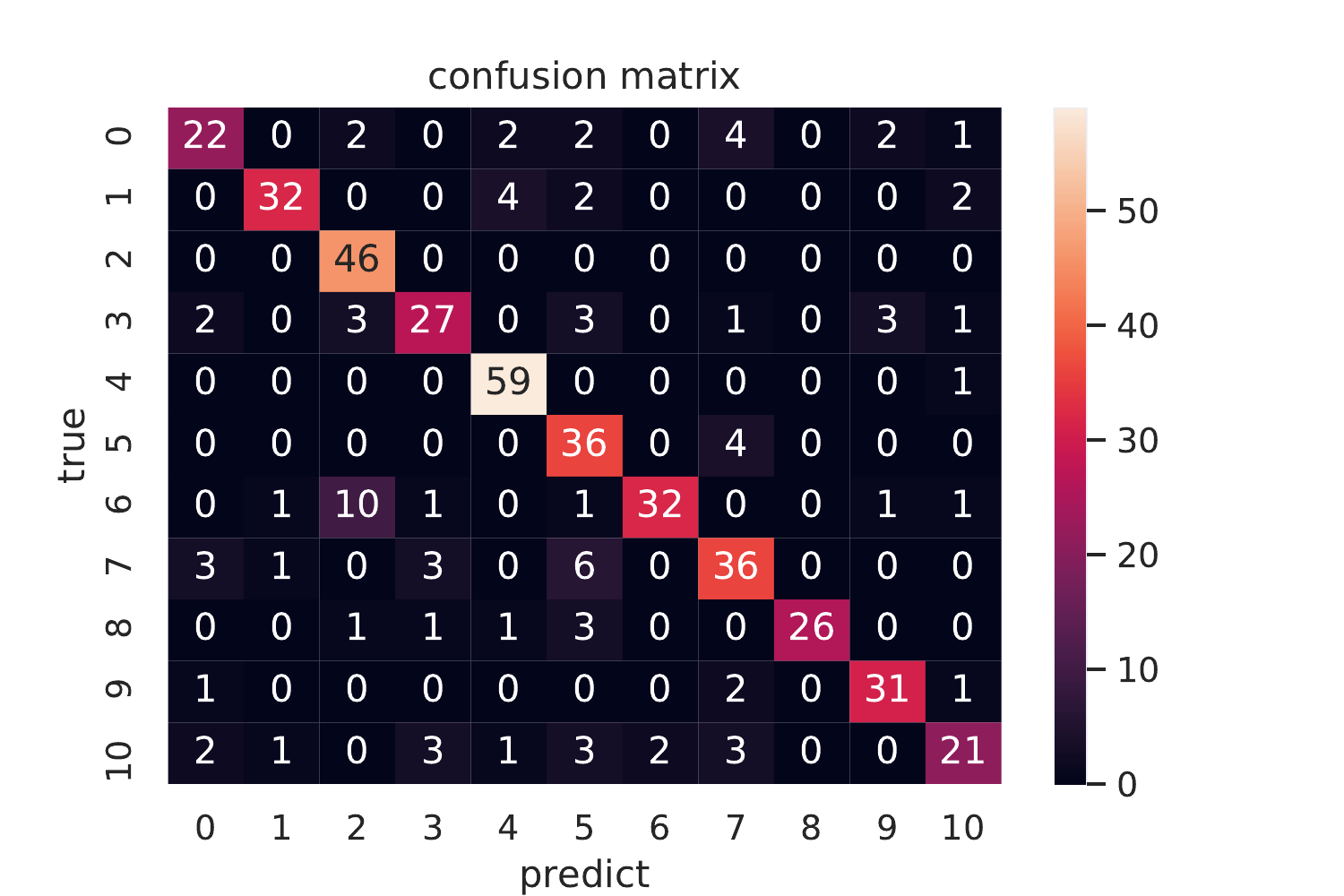}
\caption[Confusion matrix of event-based anticipation]{Confusion matrix of event-based method. Label dictionary:["Basketball":0, "Biking": 1, "Diving": 2,"GolfSwing":3, "HorseRiding":4, "SoccerJuggling":5, "Swing":6, "TennisSwing":7, "TrampolineJumping":8,"VolleyballSpiking":9,
"WalkingWithDog":10] }
\label{fig:label}
\end{figure}

\section{Conclusion}
This paper looks into the problem of using event-based data to conduct human action recognition and human action anticipation. The idea is inspired by the attribute that event-camera is sensitive to moving object in the scene.  To implement a CNN-based learning algorithm, we use the timestamp-image to encode the event data into frame-based representation and feed those timestamp-image chunk into the CNN to perform  the  action  recognition.   It  turns  out  that  a  simple  CNN-based  recognition method  can  achieve  a  competitive  performance  compares  with  those  RGB-baseline.  As for the action-anticipation task,  we use an auto-encoder to generate future event-representation  with  categorical  features,  which  enhances  the  recognition  ac-curacy when observing in-completed event sequence.  However, the performance of event-based action anticipation task does not achieve the level of the RGB baseline,and it reveals that background and appearance information are also useful in action anticipation. Future work includes exploring the potential of  RGB and event camera synthesis in human action understanding. 

\textbf{Acknowledgement}
I would like to thank Dr Miaomiao Liu for providing useful advice and GPUs for this project. I would also like to thank Mr Chaohuo Wu for helping taking the demo videos in Fig.1. 
\bibliography{references.bib}
\bibliographystyle{IEEEtran}


\end{document}